%% file: main.tex
\documentclass[conference]{IEEEtran}

\usepackage{times}
\usepackage{epsfig}
\usepackage{graphicx}
\usepackage{amsmath}
\usepackage{amssymb}

\usepackage{multirow}
\usepackage[caption=false,subrefformat=parens,labelformat=parens]{subfig}
\usepackage{booktabs,siunitx}
\usepackage{comment}
\usepackage{xcolor}

\usepackage[numbers]{natbib}
\usepackage{multicol}
\usepackage[bookmarks=true]{hyperref}

\begin{document}

%%%%%%%%% TITLE
\title{
HMPO: Human Motion Prediction in Occluded Environments for Safe Motion Planning
}

\author{\authorblockN{Jae Sung Park}
\authorblockA{Department of Computer Science\\
University of North Carolina at Chapel Hill\\
NC, USA
}
\and
\authorblockN{Dinesh Manocha}
\authorblockA{Department of Computer Science and\\Electrical \& Computer Engineering\\
University of Maryland at College Park\\
MD, USA
}}

\maketitle
\thispagestyle{empty}

\begin{abstract}

We present a novel approach to generate collision-free trajectories for a robot operating in close proximity with a human obstacle in an occluded environment. The self-occlusions of the robot can significantly reduce the accuracy of human motion prediction, and we present a novel deep learning-based prediction algorithm. Our formulation uses CNNs and LSTMs and we augment human-action datasets with synthetically generated occlusion information for training. We also present an occlusion-aware planner that uses our motion prediction algorithm to compute collision-free trajectories. We highlight performance of the overall approach (HMPO) in complex scenarios and observe upto 68\% performance improvement in motion prediction accuracy, and 38\% improvement in terms of error distance between the ground-truth and the predicted human joint positions.

\end{abstract}

\IEEEpeerreviewmaketitle

\input{1.tex}
\input{2.tex}
\input{3.tex}
\input{4.tex}
\input{5.tex}
\input{6.tex}
\input{7.tex}
\eject

\bibliographystyle{plainnat}
\bibliography{refs}

\end{document}

%% file: 1.tex
\section{Introduction}

Human motion prediction is an important part of human-robot interaction in environments where robots work in close proximity to humans.
Traditionally, industrial robots were isolated  from humans for safety.
At the same time, humans can handle jobs that require better dexterous skills than robots~\cite{lasota2015analyzing,koppula2016anticipatory}.
For some applications, it is more efficient for humans and robots to work together while sharing the same workspace. In these scenarios, it is important for a robot to observe and predict the human motion and plan its tasks accordingly.

A key challenge in achieving safety and efficiency in human-robot interaction is computing a collision-free path for the robot to reach its goal configuration.
The robot should not only complete its task but also predict the human's motion or trajectory to avoid the human as a dynamic obstacle. There is considerable work on human motion prediction as well as computation of safe trajectories.
Some recent methods predict the human motions from images or videos are based on  (CNNs)~\cite{krizhevsky2012imagenet,kataoka2016recognition,ke2016human,butepage2017deep} or Recurrent Neural Networks (RNNs)~\cite{finn2016unsupervised,fragkiadaki2015recurrent}.

When robots proximity in close proximity with humans, they gather information about the surrounding environment using visual sensors (color cameras, depth cameras, etc). Typically, head-mounted cameras on the robots observe the workspace.
As robots perform actions with their hands or arms, the moving parts of the robot may occlude the views of these  sensors.
As a result, the resulting images cannot capture information about many parts of the scenes, including the current position of the human working close to the robot~\cite{field2009motion,schick2008hand,park2017intention}.
Such occlusion by parts of a robot can prevent accurate tracking and prediction of the human motion and thereby make it hard to perform safe and collision-free motion planning.
When the robot arm occludes the input images, either the robot should determine whether the human motion can be predicted with high certainty or the robot arm should  move in such a manner that it does not occlude the field of view of the  camera (i.e. remove occlusions), as shown in Fig.~\ref{fig:robot_motion_planning1}. This results in two main challenges:

\begin{figure}[t]
  \centering
  \subfloat
  {
    \includegraphics[width=0.31\linewidth]{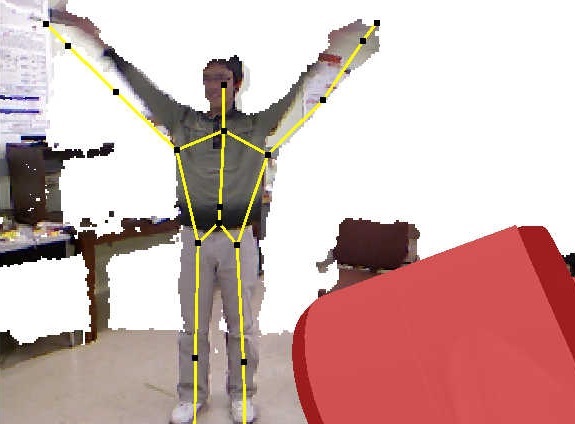}
  }
  \subfloat
  {
    \includegraphics[width=0.31\linewidth]{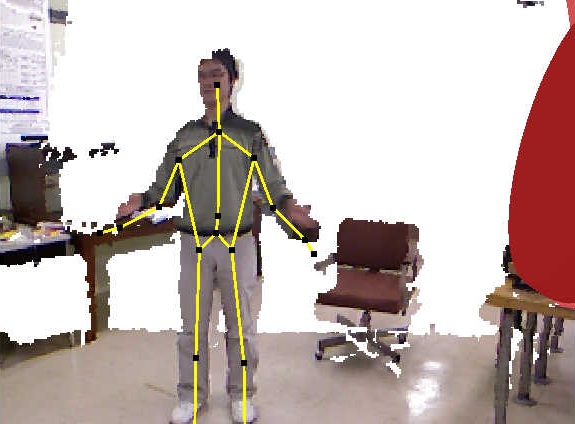}
  }
  \subfloat
  {
    \includegraphics[width=0.31\linewidth]{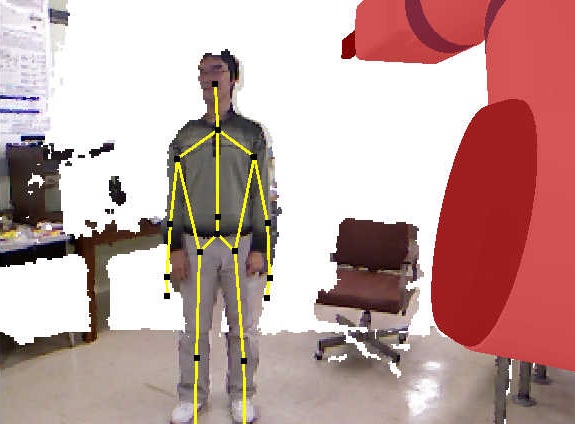}
  }
  \\
  {
    \includegraphics[width=0.9\linewidth]{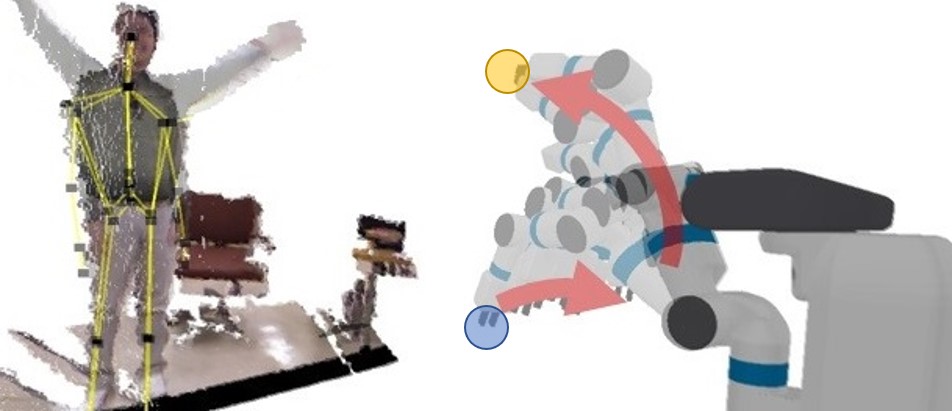}
  }
  \caption{
  A human and a robot are simultaneously operating in the same workspace. The robot arm occludes the camera view and many parts of the human obstacle are not captured by the camera. 
  Three images at the top show the point clouds corresponding to the human in the UtKinect dataset~\cite{xia2012view} for different camera positions with the occluded regions in red. The bottom right image highlights the safe motion trajectory between the initial position (blue) and the goal position (yellow). Our safe trajectory is shown in bottom right as two piece red curves (with arrows). HMPO first moves the arm to reduce the occlusion, followed by moving it to the goal position.
  }
  \label{fig:robot_motion_planning1}
\end{figure}

\begin{itemize}
\item The human motion predictor should be aware of the overlapping region between the human obstacle and the robot on the  input image. These regions occur when the human moves into the shadow region of the camera or when the robot parts occlude the region corresponding to the human in the input image. In such scenarios, prior human motion predictors do not work well.

\item The robot motion planner should respond in realtime when the human motion cannot be accurately predicted due to occlusion.
The robot motion planner should compute a safe path by taking into account these occlusion constraints.
\end{itemize}

\noindent {\bf Main Results:} We address the challenges highlighted above by presenting two novel algorithms: (1) predict human motion in the presence of obstacles and occlusion; (2) plan a robot motion, taking into account the occlusion and the certainty in the motion prediction.

  {\bf 1. Human Motion Prediction in Occluded Scenarios:} We present a neural network that uses not only the features from RGBD images, but also features related to occlusion.
Our deep learning-based approach predicts the human motion in such occluded scenarios.
We use Convolutional Neural Networks (CNNs) for feature extraction from RGBD images and feature extraction for robot occlusion.
Moreover, we use ResNet-18~\cite{He_2016_CVPR} to extract visual features from color images with occluded regions.
Our learning algorithm classifies the human action and generates the predicted human motion using a skeleton-based human model.
We add occluded images of robot scenes to existing RGBD human action prediction datasets~\cite{xia2012view,wu2015watch,dib2015}. We use these augmented datasets to train and evaluate the performance of our human motion prediction algorithm in the presence of occlusion.
In practice,  our action classification algorithm improves the prediction accuracy by $63\%$ over  prior classification algorithms~\cite{wu2015watch}.

 {\bf 2. Occlusion-Aware Motion Planning:} We present a realtime planning algorithm to compute a safe trajectory for a robot in occluded scenes with human obstacles.  
We use an optimization-based planning framework and add the occlusion constraints in the objective function. Our planner tends to compute collision-free paths and ensures that  the human region in the camera image is not occluded by the robot. We have evaluated our planner in complex environments with robots operating close to the human. In practice, our algorithm improves the overall accuracy, measured using error distance between the ground-truth and the predicted human joint positions, by $38\%$.

We use three human action RGB-D datasets and augment them with occlusion characteristics for training and validation. We highlight the performance of the overall approach (HMPO) in complex environments.
We plan to release our augmented datasets and source code at the time of publication.

%% file: 2.tex
\section{Related Work}
In this section, we give a brief overview of prior work on prediction and occlusion handling in computer vision and robotics.

\subsection{Human Motion Prediction for Robotics}

Human motion prediction has been shown to be useful to guide collaborative robots in human-robot interaction systems~\cite{unhelkar2018human}. 
The Multiple-Predictor System is a method combining multiple data-driven human motion predictors~\cite{lasota2017multiple}.
The goal-set Inverse Optimal Control algorithm plans human motion trajectories and considers them as moving obstacles in the robot motion planning step~\cite{mainprice2016goal}.
Probability models for future human motions can be used in generating collision-free robot motions.
For 2D navigation robots, the probability distribution of a human's future position on a grid map can be predicted based on a human motion model, where parameters of the motion model are approximated and learned from the motion data~\cite{fisac2018probabilistically}.
For 3D collaborative applications, the whole-body joint poses of humans may be predicted~\cite{park2017intention}.
From the tracked human skeleton joint positions, a Gaussian probability distribution can be constructed and learned through Gaussian Processes~\cite{snelson2005sparse}, and the future human motion is predicted and presented as Gaussian distributions.
All of the algorithms require fully observable information about the human motion and do not account for occlusion.
If the human motion is not fully visible, the probability distributions for non-observable human body parts will have high variances; thus the predicted future human motion is not accurate enough to generate collision-free robot motions.

\subsection{Human Motion Prediction from Images and Videos}

Motion prediction algorithms can be categorized as model-based approaches or motion analysis without an a priori human shape model~\cite{aggarwal1999human,kakadiaris1996model}.
Human motion models usually have a high degree-of-freedom (DOF) configuration space.
For skeleton model-based human models, Hidden Markov Models (HMMs) are used to predict skeleton joint positions~\cite{papadopoulos2014real}.
Deep learning-based Recurrent Neural Networks (RNNs) can be used for sequences of high-DOF human shape models~\cite{Martinez_2017_CVPR}.
An occlusion removing algorithm for self-occlusion of 3D objects and robot occlusion from robot grippers is used for robot motion planning in~\cite{hu20193}. From 3D point cloud stream data, this method recovers points that were not occluded in previous frames, but are occluded in the current frame. After recovering occluded 3D point clouds, they extract features from the point clouds and use them in RNN. However, this approach is mainly designed for deformable objects manipulated by robot grippers.
The prediction of high-DOF human motions has additional challenges due to occlusion or limited sensor ranges.
Dragan et al.~\cite{dragan2012formalizing} propose improved assistive teleoperation with predictions of the motion trajectory to reach the goal using inverse reinforcement learning.
Koppula and Saxena~\cite{koppula2016anticipating} use spatial and temporal relations of object affordances to predict future human actions.

\subsection{Object Recognition under Occlusions in a Cluttered Environment}

Self-occlusions or occlusions from surrounding objects have been investigated in the context of  object recognition and object tracking algorithms.
Multiple moving cars can be tracked from video data where some cars are occluded by others.
Without occlusions, a linear translational and scaling motion model for cars fits for tracking cars and the motions are computed by differentiating consecutive frames of images~\cite{koller1994robust}.
Prior works have also used image features to overcome the occlusion problem.
Histograms of Oriented Gradients (HOG) and Local Binary Pattern (LBP) have been considered as representative visual features and can be used in a Support Vector Machine (SVM) classifier to segment the occlusions and detect humans behind occlusions~\cite{wang2009hog} from input color images.
Human model-based body part tracking under an occluding blanket in hospital monitoring applications has been developed~\cite{achilles2016patient}. This is a specialized technique for this application.
From input depth images of a human occluded by obstacles~\cite{dib2015}, human joint positions can be tracked from a hierarchical particle filter, where occlusions are handled with a 3D occupancy grid and a Hidden Markov Model (HMM) is used to represent the state of visibility and occlusion. However, it is unable to track parts that are not visible.
To overcome and respond to occlusions in object recognition or human body pose estimation, the visibility of occluded objects or human body parts can be computed using supervised learning~\cite{haque2016towards,martinez2017object}.
By labeling the visibility of body parts with 0 and 1 in the training data and minimizing the loss function for visibility, the visibility is then inferred as a probability in the range of [0, 1].

Our approach is more general and complimentary than the methods discussed above. Not only do we present a novel deep learning-based method to predict human motion in occluded scenarios, but we also compute a motion trajectory for a robot that reduces occlusions. Moreover, we exploit robot kinematics and self-occlusion capabilities to achieve higher classification accuracy than prior methods.

%% file: 3.tex
\section{Overview}
In this section, we describe our problem and the assumptions made by our algorithm. Furthermore, we give an overview of the overall approach combining human motion prediction and occlusion-aware motion planning.
\subsection{Problem Statement and Assumptions}

\begin{figure*}[t]
    \centering
    \includegraphics[width=\linewidth]{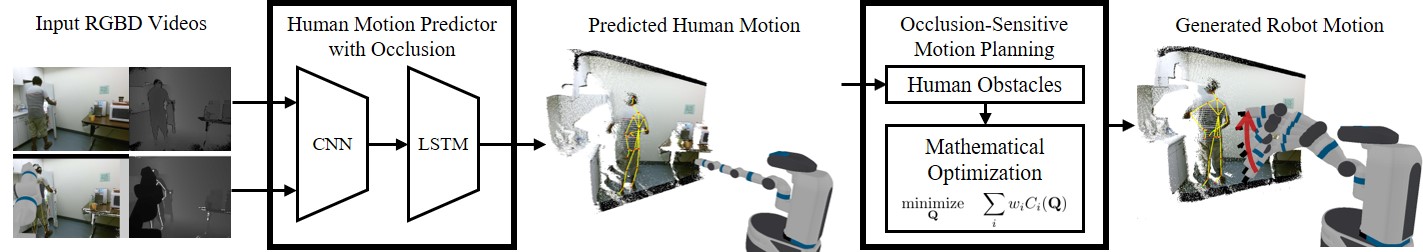}
    \caption{ {\bf HMPO:}
    Overall pipeline of our human motion prediction and robot motion planning. We present a new deep learning technique for human motion prediction in occluded scenarios and an optimization-based planning algorithm that accounts for occlusion.
    }
    \label{fig:overview}
\end{figure*}

Figure~\ref{fig:overview} highlights the different components of our approach.
In our environment, we assume that there is a collaborative robot with one or more robot arms and a camera. Moreover, the robot is operating in close proximity to a human obstacle, and our goal is to compute a collision-free and safe trajectory for the robot. We assume that the human is active and the robot is passive while the robot arm shares the same workspace with the human.
The human either performs actions as if there were no robots nearby or as if he or she believes the robot will avoid collisions.

In these scenarios, the robot tracks and predicts the motion of the human using the camera and uses that information for safe planning. We extract the human skeleton from the image and uses the skeleton for motion prediction (see Fig.~\ref{fig:robot_motion_planning1}).
Our approach is designed for environments, where the robot's motion results in self-occlusions with respect to the camera. This happens for configurations, where the robot arm either fully or partially occludes the human.
The input of the human motion predictor is captured from a single RGBD camera attached to the robot's head. Our approach can also work with 2D RGB cameras.  
The RGB and depth image frames are fed as input to the human motion predictor at a fixed frame rate, which is governed by the underlying camera hardware and the training datasets.
For example, the Kinect V2 sensor streams color and depth images at $30$ frames per second.
The camera position and angle are set to capture the human's motion.
The outputs of the human motion predictor are the human action, the future human motion with the skeleton-based human model, and the certainty value related to the probability that the human motion can be predicted accurately in the occluded scenarios.

\noindent {\bf Real-time Planning:} We  present an occlusion-aware realtime motion planning algorithm. Our planner takes as input the current configuration of the robot, including the arm, and computes a high-dimensional trajectory in the configuration space that is represented in the space corresponding to the  robot configuration $\mathbf{q} \in \rm I\!R^n$ and the time $t \in \rm I\!R$.
The trajectory connects the robot's  configuration at the current time to the goal configuration at a later time.
The future motion of the human is predicted from our deep learning-based human motion predictor, represented using a skeleton-based model. Our planner takes this predicted trajectory into account for safe motion planning. Our planner modifies this trajectory in real-time in response to the obstacles in the environment  and considers two constraints:
\begin{enumerate}
    \item Collision avoidance with static obstacles and predicted paths of dynamic obstacles, especially humans.
    \item Moving the robot arms so they do not occlude the human from the camera's point of view. This way, the accuracy of the human motion predictor will improve in subsequent frames.
\end{enumerate}
 We present an optimization-based planner based on these constraints.

%% file: 4.tex
\section{Human Motion Prediction with Occluded Videos}

In this section, we present our novel human motion prediction algorithm that accounts for occlusions in the scene.
\subsection{Neural Network for Occluded Videos}
\label{subsec:neural_network}

\begin{figure*}[ht]
  \centering
  \includegraphics[width=0.95\linewidth]{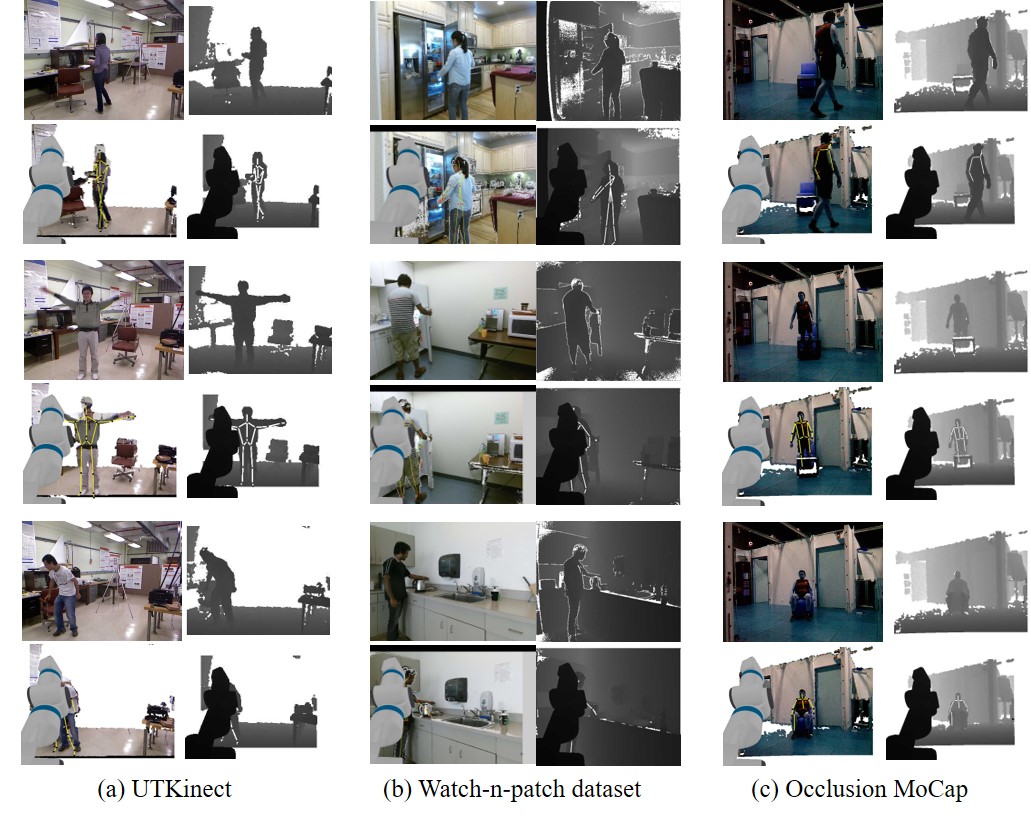}
  \caption{
  Sample images of original datasets and modifications with occlusion information. (a) UTKinect dataset~\cite{xia2012view}. (b) Watch-n-patch dataset~\cite{wu2015watch}. (c) Occlusion MoCap dataset~\cite{dib2015}.
  We present $3$ image pairs for each dataset in each column. The top image in each pair is the original image from the dataset, and the bottom images are generated by augmenting the original images with robot arm occlusions at the bottom. These augmented images are used for training and cross-validation.
  }
  \label{fig:dataset}
\end{figure*}

Our approach is based on convolutional neural networks (CNNs), which have been widely used for image classification and recognition~\cite{krizhevsky2012imagenet,kataoka2016recognition,ke2016human,butepage2017deep}.
We first extract the features, which are used by LSTMs, from the pre-trained ImageNet~\cite{krizhevsky2012imagenet}. In addition to the image features, we also take into account occlusion features.
The deep neural network is provided with the input color image sequence, the depth image sequence, and an occlusion mask image sequence.
To facilitate the robot's early response, we need to predict the human action class quickly.

The input image sequence contains the human upper body action. The color and depth images may be occluded by the robot arm, and it is assumed that the robot knows which parts of the images are being occluded, as shown in the red regions in Figure~\ref{fig:robot_motion_planning1}. We use forward kinematics based on robot joint values and the robot camera position to compute the occlusion region in the image. 
The output corresponds to the human action class, the future human motion in a short time window, and the confidence value of the human motion prediction. For action classification, our prediction algorithm outputs a discrete probability distribution for various action classes included in the datasets. For the future human motion, the human skeletal joint positions are predicted. Those predictions will have a 100\% confidence level, if the robot's configuration does not result in self-occlusions. The confidence level decreases when the human motion is partially occluded; at 0\%, the human motion is completely hidden.

Recurrent Neural Networks and Long Short-Term Memory (LSTM) models are useful for constructing deep neural networks for temporal sequences. We exploit these models to predict human actions and future motions with the RGBD input image sequences, which may be partially occluded by the robot arm. In addition to the pre-trained CNN features from the color and depth images, we also use a neural network input for the occlusion image to adjust the human motion prediction results and generate the confidence level of the certainty with which the human motion can be predicted. The feature vectors of color and depth and the occlusion images are fed to the LSTM.
The features from depth images and occlusion images are different and are used to generate accurate confidence level results.
The output contains the information about action classification, future human joint position, and degrees of occlusions.
For each action class, a real value between 0 and 1 represents the likelihood that the human is performing a certain action.
The predicted action is the one with the highest value among the action classes.

The input color and depth images are first cropped around the human with the resolution $224 \times 224$ to feed the input for resnet-18. The output of the pre-trained CNN is a vector of size $1,000$ for each color and depth image. The column vector describing the skeleton joint positions has $x$, $y$, $z$ components for each joint. These values are concatenated and connected to a fully-connected layer of size $1,000$ followed by LSTM.

The outputs of the neural network are the $x$, $y$, and $z$ components of the future human joint position, future human action class, and the confidence value.
Future human joint positions are predicted up to $3$ seconds ahead of time. The 3-second time window is discretized using $0.5$s timesteps (i.e. ``prediction timestep''), resulting in $6$ time points at which the joint positions are predicted. The x, y, and z coordinates of each joint compose the output vector. The degree of occlusion is represented by a real value between $0$ and $1$. A value near $1$ implies that the joint position is difficult to predict due to robot occlusions, whereas a value near $0$ means the joint is not occluded by the robot. To train the future joint positions, the ground-truth joint positions in the sequence for each timestep ahead of the current time are used as the expected outputs.
To avoid the redundancy of temporal relationships from LSTM and the output values, the values for predicted joint positions and the degrees of occlusion in the output layer do not interconnect with those values from different time points.

\subsection{Dataset Generation}
\label{subsec:dataset_generation}

In the field of computer vision, synthetic data has been used widely, reducing the efforts of collecting data and improving prediction performance~\cite{song2017semantic,varol2017learning}. There is very little data from real-world scenarios in terms of humans reacting when they are close to robots. Usually, when robot motion planners work in close proximity with humans in the real-world, the color and depth cameras are installed at a location that minimizes the robot occlusions and human self-occlusions while still accurately tracking human skeleton joints. As a result, synthetic datasets are used to generate results for our supervised learning method. Our synthetic datasets have robot images overlaid on the original dataset, as if the robot arm image was captured from the viewpoint of the head-mounted camera.

To train the neural network, we extend three existing datasets for training and cross-validation by adding robot occlusions in the images.
There may be some small errors in synthesized datasets, such as pixel color values, depth values, and joint angles of actual motors, compared to real-world captured images with robot occlusions. However, our main problem is predicting the human joint positions and human action class behind the robot occlusion, and the regions of occlusion from forward kinematics. Our approach provides a robust solution to predict human motions accurately with synthesized training data.
Furthermore, we added a new action class in these datasets to represent whether the human is occluded by the robot.

\noindent \textbf{UTKinect-Action} dataset~\cite{xia2012view} (Figure~\ref{fig:dataset} (a)) contains 10 types of human actions (\textit{Walk}, \textit{Sit Down}, etc.) and each action has about $18$ to $20$ RGBD videos captured with Kinect v1. The resolution of the RGB videos is $640 \times 480$, whereas the resolution of the depth videos is $320 \times 240$. The actions are performed with $10$ different subjects. The videos are captured in the same space (a lab) with the same Kinect position and angle.

\noindent \textbf{Watch-n-Patch} dataset~\cite{wu2015watch} (Figure~\ref{fig:dataset} (b)) provides RGBD videos of $21$ types of human actions performed by $7$ subjects captured with Kinect v2. The resolution of the RGB videos is $1920 \times 1080$, whereas the resolution of the depth videos is $512 \times 424$. The videos are captured in $8$ offices and $5$ kitchens with different Kinect positions and angles.

\noindent \textbf{Occlusion MoCap} dataset~\cite{dib2015} (Figure~\ref{fig:dataset} (c)) has RGBD videos of a human with joint tracking Qualysis markers on his body and a static object in the middle of the room. There are 4 videos with lengths between $45$ and $60$ seconds captured at $15$ frames per second. In the videos, a person comes into the space, walks around the chair in the middle of the space, and sits down. The dataset has $640\times480$ resolution in both color and depth images. While the action labels are not given in the dataset, this one provides more accurate joint positions than the other two datasets highlighted above.

In all the datasets, only one human subject performs the actions and human skeleton tracking data are available.
We add a robot arm occlusion in both the RGB videos and depth videos of the UTKinect, Watch-n-Patch, and Occlusion MoCap datasets to make them effective for our prediction algorithm.
The robot occlusions are added as if the videos are captured by a camera on a virtual robot, where the robot arm is moving around in the same space that is used to perform human actions.
The inserted robot occlusions are rendered with simulated geometric models of the robot and appropriate models of light to simulate the images and occlusion. The regions of occlusion are computed using forward kinematics. It is accurate up to the resolution of the image-based methods.
Because the humans in the original dataset are moving without the presence of robot, those captured human motions are neither changed nor affected in the occluded datasets. Therefore, the virtual robot's goal is to avoid collisions with the humans. In order to generate the virtual robot's motion, we used the ITOMP optimization-based motion planner~\cite{Park:2012:ICAPS} to avoid collisions along with probabilistic collision detection~\cite{1902.10252} to measure collision probability with noisy point cloud data.

The file sizes of the UTKinect, Watch-n-Patch, and Occlusion MoCap datasets are 7GB, 30GB, and 2GB, respectively, and we generate additional input images with occlusions.
Duplicating image files and saving them in storage disks can be inefficient, so we store the synthesized dataset by only storing the robot joint angles for each frame.
From the robot joint poses, the RGBD images and occlusion images are obtained by overlaying the robot image on the original images.

When human motions are not fully visible due to occlusions, human action labels cannot be predicted accurately. In this case, we  semi-automatically assign an \textit{occluded} label.
To determine if the human action can be predicted, we check if the human skeleton tracking data is occluded by the generated virtual robot arm motions.
For action labels that are recognized mostly from human hand motions (e.g., \textit{fetch-from-fridge}, \textit{drinking}, or \textit{pouring}), the human action cannot be predicted if the robot arm occludes the human hand.
These action labels are changed to \textit{occluded} if the human hand joint is occluded by the virtual robot in the depth image.
For other action labels that are recognized from the motion of the whole body (e.g., \textit{walking}, \textit{leave-office}, or \textit{leave-kitchen}), the human action can be predicted if some parts in the RGBD videos are occluded but cannot be predicted if most parts of the human are occluded.
These action labels are changed to \textit{occluded} if most of the human joints are occluded by the virtual robot.
There are $23$ joints in the human skeleton tracking data.
We label \textit{occluded} if 20 or more joints are occluded.
For the prediction algorithm to be able to predict actions when RGBD videos are not occluded, the original datasets are also included in the training dataset without modification.

The neural network is given the images with occlusions for both training and inference. The synthesized datasets include images without robot occlusions when the robot arm does not occlude the camera. About 50\% of the training dataset images have robot occlusions to train human action and joint positions behind occlusions. These data have the \textit{occluded} label and a 0 confidence value for expected output if the robot parts occlude more than half of the human joints. The rest of the images with no occlusions are also necessary to train human action and joint positions without occlusions. These data with and without robot occlusions would be used in the real-world scenarios.
The human motion prediction and occlusion-aware motion planner work well without occlusion because the training dataset contains images without occlusions. The robot occlusion does not hide the human, where the certainty values are 1 and the robot motion trajectory is not affected by occlusion-related cost functions. The algorithms also work well with occlusion.

%% file: 5.tex
\section{Occlusion-Aware  Motion Planning}
In this section, we describe our planning algorithm that uses the human motion prediction results computed in the prior section.

\subsection{Optimization-Based Planning of Robot Trajectories}

We denote a single configuration of the robot as a vector $\mathbf{q}$, which consists of joint-angles or other degrees-of-freedom.
An n-dimensional configuration at time $t$, where $t \in \mathbb{R}$, is denoted as $\mathbf{q}(t)$. We assume $\mathbf{q}(t)$ is twice differentiable, and its first and second derivatives are denoted as $\mathbf{q}'(t)$ and $\mathbf{q}''(t)$, respectively.
We represent bounding boxes of each link of the robot as $B_i$. The bounding boxes at a configuration $\mathbf{q}$ are denoted as $B_i(\mathbf{q})$.

For a planning task with the given start configuration $\mathbf{q}_s$ and goal configuration $\mathbf{q}_g$, the robot's trajectory is represented by a matrix $\mathbf{Q}$, the elements of which correspond to the waypoints~\cite{STOMP:2011,zucker2013chomp,Park:2012:ICAPS}:
\begin{equation}
\mathbf{Q} = \begin{bmatrix}
    \mathbf{q}_0 & \mathbf{q}_1 & & \mathbf{q}_{n-1} & \mathbf{q}_n \\
    \mathbf{q}'_0 & \mathbf{q}'_1 & \cdots & \mathbf{q}'_{n-1} & \mathbf{q}'_n \\
    t_0 = 0 & t_1 & & t_{n-1} & t_n = T
\end{bmatrix}.
\end{equation}
The robot trajectory passes through $n+1$ waypoints $q_{0}, \cdots, q_{n}$, which will be optimized by an objective function under constraints in the motion planning formulation.
Robot configuration at time $t$ is cubically interpolated from two waypoints.

We use optimization-based robot motion planning~\cite{Park:2012:ICAPS} for generating robot trajectories in dynamic scenes.
The objective function for the optimization-based robot motion planning consists of different types of cost functions.
The $i$-th cost functions of the motion planner are $C_i(\mathbf{Q})$.
\begin{equation}
\begin{aligned}
& \underset{\mathbf{Q}}{\text{minimize}}
& & \sum_i w_i C_i(\mathbf{Q}) \\ \label{optimize}
& \text{subject to}
& &
\begin{array}{l}
  \mathbf{q}_{min} \leq \mathbf{q}(t) \leq \mathbf{q}_{max}, \\
  \mathbf{q}'_{min} \leq \mathbf{q}'(t) \leq \mathbf{q}'_{max}, \\
  \mathbf{q}_{0} = \mathbf{q}_{s}, \quad \mathbf{q}_{n} = \mathbf{q}_{g}
\end{array}
\, \, 0 \leq \forall t \leq T,
\end{aligned}
\vspace{-6pt}
\end{equation}
for the initial robot configuration $\mathbf{q}_{s}$ and the goal configuration $\mathbf{q}_{g}$.
In the optimization formulation, $C_i$ is the $i$-th cost function and $w_i$ is the weight of the cost function.
Every $0.5$s timestep, the motion planning problem is updated, and the motion planner adjusts the trajectory with respect to changes in human motions and prediction of occlusion and human action.

In a static environment where there are no humans or dynamic obstacles, we define the basic cost functions: robot smoothness and collision avoidance with static obstacles.

\emph{\bf Smoothness:}
\begin{align}
C_{smoothness}(\mathbf{Q}) = \frac{1}{T} \int_0^T \mathbf{q}'(t)^T \mathbf{D} \mathbf{q}'(t) dt,
\end{align}
where $\mathbf{D}$ is a diagonal matrix with non-negative values.

\emph{\bf Collision avoidance with static obstacles:}
\begin{align}
C_{collision}(\mathbf{Q}) = \frac{1}{T} \int_0^T \sum_i \sum_j \textit{dist}(B_i(t), O_j)^2 dt,
\end{align}
where $\textit{dist}(B_i(t), O_j)$ is the penetration depth between a robot bounding box $B_i(t)$ and a static obstacle $O_j$.

\subsection{Occlusion Sensitive Constraints}
\label{subsec:robot_motion_planning}

We account for occlusion characteristics by adding  a new soft constraint that prevents the robot from occluding the human obstacle, especially when the certainty in motion prediction is low.

\emph{\bf Robot occlusion:}
\begin{align}
C_{occlusion}(\mathbf{Q}) = \frac{1}{T} \int_0^T (1 - \alpha(t))^2 dt, \label{eq:robot_occlusion}
\end{align}
where $\alpha(t)$ is the confidence value at time $t$ of human motion prediction, where the robot may have occluded the human image captured by the RGBD sensor.
The confidence value is one of the output values of the neural network in Section~\ref{subsec:neural_network} and is in the range $[0, 1]$.
A confidence value near $1$ means that the human is not very occluded by the robot, whereas a value near $0$ means that the human motion cannot be accurately predicted.
We modify the trajectory to reduce $C_{occlusion}$ and this reduces the overlapping area of the robot and the human portion in the RGBD frames over the duration of the trajectory.

\subsection{Real-time Collision Avoidance with Predicted Human Motions}

In order to avoid collisions with the human obstacle in the 3-second future time period, we add a soft constraint that imposes a penalty in terms of the extent of the penetration depth between the robot and the predicted human motion.

\emph{\bf Collision avoidance with a human:}
\begin{align}
C_{collision}(\mathbf{Q}) = \frac{1}{T} \int_0^T \sum_i \sum_j \textit{dist}(B_i(t), H_j(t))^2 dt, \label{eq:collision_with_human}
\end{align}
where $\textit{dist}(B_i(t), H(t)_j)$ is the penetration depth between a robot bounding box $B_i(t)$ and the predicted human obstacle $H_j(t)$ at time $t$.
The human obstacle is represented with multiple capsules, each of which connects a pair of joints.
$H_j(t)$ represents a capsule with index $j$, connecting two human joints $\mathbf{h}_{j,1}(t)$ and $\mathbf{h}_{j,2}(t)$, where the joint positions come from the result of the skeleton model-based human motion prediction in Section~\ref{subsec:neural_network}
For the prediction uncertainty of each joint due to the presence of occlusions, we change the radius of the capsule with respect to the confidence values for the joints $\alpha_{j,1}(t)$ and $\alpha_{j,2}(t)$.
To reduce the computation time, we take the average of two confidence values and the radius $r_j(t)$ is linearly interpolated as:
\begin{align}
    \alpha_j(t) &= \frac{1}{2} (\alpha_{j,1}(t) + \alpha_{j,2}(t)), \\
    r_j(t) &= (1 - \alpha_j(t)) r_0 + \alpha_j(t) r_1, \quad r_0 \geq r_1
\end{align}
where $r_0$ and $r_1$ are user-specified parameters.
When the occlusion confidence $\alpha_j(t)$ is $0$, this implies that the joints are occluded and the radius is $r_0$.
On the other hand, when $\alpha_j(t)$ is $1$ that implies that the joints are not occluded, and the radius is $r_1$.

%% file: 6.tex
\section{Performance and Analysis}

\subsection{Human Action Recognition and  Motion Prediction}

After generating RGB-D datasets with occlusion characteristics (see Section~\ref{subsec:dataset_generation}), we use them for training and evaluation. 
The Watch-n-patch dataset~\cite{wu2015watch} has a frame rate of 5 frames per second.
Each dataset has two types of RGB-D images: \textit{No Occlusion} and \textit{Occlusion} (see Fig.~\ref{fig:dataset}). We perform  5-fold cross-validation, and these datasets are divided into $5$ segments.
$4$ segments are used for training and the remaining one is used for validation.
When splitting the dataset, we split the original dataset into $5$ subsamples, and we split the modified dataset with robot occlusions into $5$ subsamples.
4 subsamples of the original dataset and 4 subsamples of the modified dataset are used for training, and the remaining subsamples are used for validation.

We have tested our neural network models by enabling and disabling the input data channels related to the robot occlusion.
These input channels are: \textit{Occlusion Color}, \textit{Occlusion Depth}, and \textit{Skeleton}.
\textit{Occlusion Color} is the color image of the robot with a white background.
\textit{Occlusion Depth} is the depth image of the robot with a white background.
\textit{Skeleton} is the tracked human skeletal joint positions in 3D coordinates with respect to the camera coordinate system.
The baseline planning algorithm only accepts the color and depth images and does not acquire information about robot occlusions.
We created $7$ different models or versions of planners by enabling the three input channels described above.
HMPO accepts color image, depth image, color robot occlusion image, depth robot occlusion image, and the tracked human skeleton.

\begin{figure*}[t]
  \centering
  \subfloat
  {
    \includegraphics[width=0.15\linewidth]{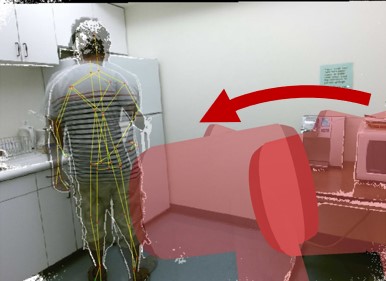}
  }
  \subfloat
  {
    \includegraphics[width=0.15\linewidth]{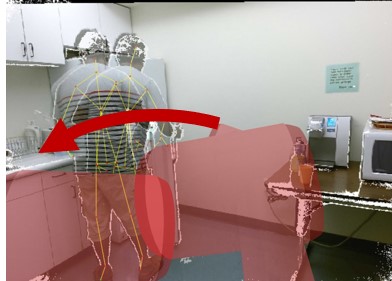}
  }
  \subfloat
  {
    \includegraphics[width=0.15\linewidth]{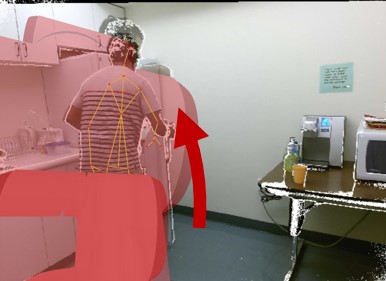}
  }
  \subfloat
  {
    \includegraphics[width=0.15\linewidth]{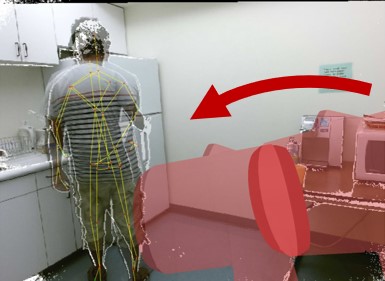}
  }
  \subfloat
  {
    \includegraphics[width=0.15\linewidth]{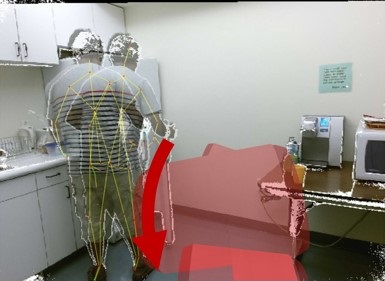}
  }
  \subfloat
  {
    \includegraphics[width=0.15\linewidth]{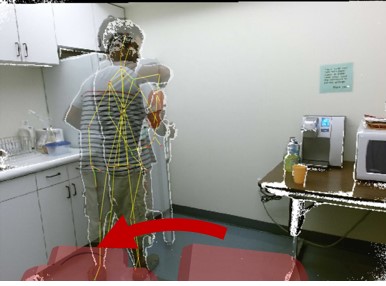}
  }
  \\
  \addtocounter{subfigure}{-6}
  \subfloat[]
  {
    \includegraphics[trim=0 30 0 10,clip,width=0.45\linewidth]{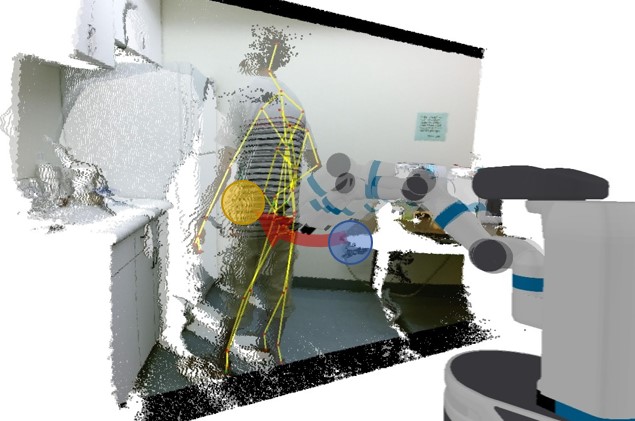}
  }
  \subfloat[]
  {
    \includegraphics[trim=0 30 0 10,clip,width=0.45\linewidth]{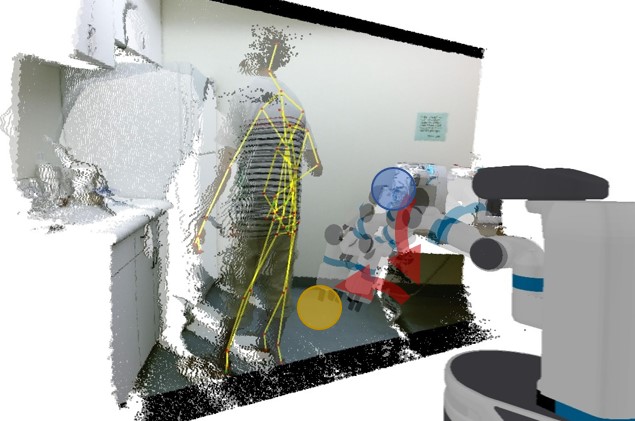}
  }
  \caption{
  {\bf Benefits of Occlusion-Aware Planning:} The top row highlights the point cloud with the dynamic human obstacle, and the regions occluded by robot arms (in red). The bottom row highlights the trajectories computed by different planners when as the robot arm needs to move from right to left:
  (a) The trajectory is generated by the baseline planner, which does not account for occlusion.
  When the robot occludes the human,  the motion prediction error is high and results in collisions.
  (b) The robot arm motion is generated by our occlusion-sensitive planner. The arm first moves to reduce the level of occlusion (i.e. a detour) and then reaches the goal to compute a safe trajectory.
  }
  \label{fig:robot_motion_planning2}
\end{figure*}

\begin{figure*}[ht]
  \centering
  \includegraphics[width=0.95\linewidth]{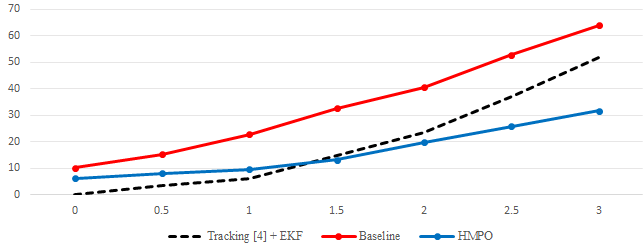}
  \caption{
  Average error distance over time for up to $3$ seconds between ground truth joint positions and the predicted joint positions. The error distances for the skeleton tracking~\cite{dib2015} and Extended Kalman Filter (EKF) are shown with a dashed line, and the error distances for our models are shown with solid lines. The baseline model without occlusion input images has higher error distances. However, the HMPO model has better prediction results with lower error distance values than EKF when the future prediction time is 1.5 seconds or higher.
  }
  \label{fig:error_distance}
\end{figure*}

\begingroup
\begin{table*}[t]
  \centering
  \begin{tabular}
  {
    cccc
  }
  \toprule
  \multirow{2}{*}{Error Distance (cm)} &
        \multirow{2}{*}{UTKinect~\cite{xia2012view}} &
        \multirow{2}{*}{Watch-n-Patch~\cite{wu2015watch}} &
        \multirow{2}{*}{Occlusion MoCap~\cite{dib2015}} \\
    & & & \\ \midrule
    Tracking~\cite{dib2015} + EKF &  &  & 51.6 (17.7) \\ \midrule
    Baseline                      & 91.3 (26.8) & 116 (28.4) & 64.0 (16.7) \\
    Occlusion Color               & 94.1 (20.4) & 110 (22.9) & 63.4 (14.5) \\
    Occlusion Depth               & 83.1 (21.6) & 105 (28.2) & 41.0 (9.3) \\
    Skeleton                      & 79.9 (15.2) & 96.8 (19.7) & 38.6 (9.2) \\
    Occlusion Color + Depth       & 72.9 (15.0) & 91.4 (21.4) & 35.4 (14.9) \\
    Occlusion Color + Skeleton    & 70.9 (13.0) & 82.7 (21.4) & 34.0 (4.9) \\
    Occlusion Depth + Skeleton    & 65.3 (12.1) & 77.1 (22.7) & 35.1 (4.0) \\
    HMPO                    & 61.9 (15.8) & 76.8 (14.3) & 31.8 (6.9) \\
    \bottomrule
  \end{tabular}
  \caption{{\bf Accuracy Comparison of Prediction Algorithms on  Different Datasets:} Average error distance (lower is better) between ground truth joint positions and the predicted joint positions after $3$ second for different datasets and algorithms.
  The numbers in parentheses are standard deviations.
  The baseline is based on tracking methods~\cite{dib2015}  along with extended Kalman filters on the skeleton-based human motion model.
  Our approach, HMPO (31.8 cm), reduces the error distance dataset by $38\%$ from the particle filter-based tracking~\cite{dib2015} plus Extended Kalman Filter (51.6 cm) and $50\%$ from the baseline (64.0 cm). This demonstrates the accuracy benefits of our occlusion-aware planner.
  }
  \label{table:accuracy_joint}
\end{table*}
\endgroup

We measure the performance of our joint position prediction and action classification algorithms.
Table~\ref{table:accuracy_joint} shows the performance of the future human joint position prediction for the different classification models.
The average {\em error distance} is measured as follows:
\begin{align}
    d_{err}(t) = \frac{1}{N} \sum_{i=1}^N || \mathbf{h}_{i}(t) - \mathbf{h}_{truth,i}(t) ||,
\end{align}
where $N$ is the number of human skeleton joints, $\mathbf{h}_{i}(t)$ is the predicted $i$-th human 3D joint position at time $t$, and $\mathbf{h}_{truth,i}(t)$ is the ground-truth human joint position.
The human skeleton model-based joint tracking with particle filter~\cite{dib2015} has an average error distance of 16.0 cm for tracking.
An Extended Kalman Filter with linear motion of joint angles is used to predict the future joint positions.
With the particle filter and the Extended Kalman Filter, the average prediction error is 34.0 cm, which is a significant increase over the average tracking error of 16.0 cm.
When occlusion characteristics are added to the RGB-D images, the error distance increases to $51.6$ cm.
The error distance of HMPO in the \textit{Occlusion} dataset is 31.8 cm.
HMPO reduces the error distance dataset by $38\%$ from the particle filter-based tracking~\cite{dib2015} plus Extended Kalman Filter (51.6 cm) and $50\%$ from the baseline (64.0 cm).

\begin{table}
  \centering
  \begin{tabular}
  {
    @{}
    cc
    @{}
  }
  \toprule
    \multirow{2}{*}{Accuracy (\%)} & Dataset \\ \cmidrule{2-2}
    & Watch-n-Patch~\cite{wu2015watch} \\ \midrule
    Wu et al.~\cite{wu2015watch} & 22.5 \\ \midrule
    Baseline                   & 19.7 (6.3) \\
    Occlusion Color            & 16.9 (5.0) \\
    Occlusion Depth            & 24.4 (5.2) \\
    Skeleton                   & 28.8 (6.1) \\
    Occlusion Color + Depth    & 28.3 (4.3) \\
    Occlusion Color + Skeleton & 30.7 (7.1) \\
    Occlusion Depth + Skeleton & 31.0 (5.4) \\
    HMPO                 & 36.6 (4.1) \\
    \bottomrule
  \end{tabular}
  \caption{Accuracy of action classification and human motion prediction algorithms for the Watch-n-Patch dataset (higher is better).
  The numbers in parentheses are standard deviations.
  {
 HMPO ($36.6\%$) improves the action classification accuracy in the \textit{Occlusion} dataset by $63\%$ from Wu et al.~\cite{wu2015watch} ($22.5\%$) and $86\%$ from the baseline ($19.7\%$).
  }
  }
  \label{table:accuracy_class}
\end{table}

Table~\ref{table:accuracy_class} highlights the performance of  human action class prediction for different classification models.
Wu et al.~\cite{wu2015watch} highlighted $31.6\%$ accuracy on action classification for the original Watch-n-patch dataset with $21$ different types of human action classes.
When robot occlusion is added to this dataset, human skeleton-based visual features cannot be extracted. This results in lower accuracy of classification ($22.5\%$) for both the original action class labels and the \textit{occluded} label.
However, when more input channels containing information about occlusions are added to the baseline, the classification accuracy increases.
We observe that \textit{Occlusion Depth} and \textit{Skeleton} inputs play a more significant role in terms of action classification for the \textit{Occlusion} dataset than \textit{Occlusion Color}.
Overall, the accuracies of the \textit{Occlusion Depth} and \textit{Skeleton} for \textit{Occlusion} datasets increase from the accuracy of the baseline ($19.7\%$) by $4.7pp$ and $9.1pp$, respectively.
However, the accuracy of \textit{Occlusion Color} decreases by $2.8pp$ from the baseline, though the occlusion color input channel contributes to an increase when combined with the occlusion depth or the skeleton input channels.
The classification accuracy of HMPO is $36.6\%$.
HMPO improves the action classification accuracy in the \textit{Occlusion} dataset by $63\%$ from Wu et al.~\cite{wu2015watch} ($22.5\%$) and $86\%$ from the baseline ($19.7\%$). This demonstrates the benefits of our approach.

\begin{comment}
\begin{figure*}[ht]
  \centering
  \includegraphics[width=0.95\linewidth]{figs/dummy.png}
  \caption{
  Sample images of human motion prediction under occlusion and robot trajectories.
  }
  \label{fig:robot_motion_planning3}
\end{figure*}
\end{comment}

\subsection{Occlusion-aware Motion Planning}

We use the Fetch robot with an RGB-D camera on its head and a 7-DOF robot arm.
The environments are represented as point clouds of human and static objects from the RGB-D datasets.
In addition, we add virtual tables and bookshelves to the environments, so that the robot can interact with them as static obstacles.
The robot's task is to move a simple object on the table or bookshelf to a goal location while avoiding collisions with static obstacles and the human (see Fig.~\ref{fig:robot_motion_planning2}).
The initial and goal locations of the object are randomly set for each task.
The moving task is repeated with randomized goal locations for our evaluations.

The human joint positions occluded by the robot arm are set to zero (untracked) as they are used as inputs to the LSTM described in Section~\ref{subsec:neural_network}. Only the inferred future joint positions and the confidence values are used while computing the collision and occlusion cost functions in our planner.
To evaluate the performance, robot motion trajectories are generated from a baseline  planner without the robot occlusion cost functions (left) and from our occlusion-aware robot motion planner, which uses the robot occlusion cost function (right) in Figures~\ref{fig:robot_motion_planning1} and ~\ref{fig:robot_motion_planning2}, respectively.
The baseline robot motion planner tends to generate trajectories that collide with the human when the robot arm occludes the human from the robot head camera in the input images.  This demonstrates the benefits of our planner, as it is able to compute a collision-free path in a complex environment with occluded dynamic obstacles.
\begin{comment}
Figure~\ref{fig:robot_motion_planning3} shows sample images of prediction results and robot trajectories.
\end{comment}

%% file: 7.tex
\section{Conclusion and Limitations}

We present a novel approach to generating safe and collision-free trajectories for a robot operating in close proximity with a human obstacle. In these scenarios, parts of the robot (e.g., the arms) can result in self-occlusion and reduce the accuracy of human motion prediction. We present two novel algorithms. The first of these is a deep learning-based method for human motion prediction in occluded scenarios that not only considers image features but also occlusion features for training and evaluation. We use three widely used datasets of human actions and augment them with synthetic occlusion information. Compared to prior classification algorithms, we observe up to $68\%$ improvement in motion prediction accuracy. Second, we present an occlusion-aware planner that considers the predicted trajectories and the confidence level. It directly computes a safe trajectory or moves the robot arms to reduce the extent of occlusion, thereby increasing the accuracy of human motion prediction for safe planning. We have highlighted the performance in complex scenarios where prior planners are unable to compute collision-free trajectories. Furthermore, we observe up to $38\%$ improvement in terms of the error distance metric. To the best of our knowledge, this is the first general method for safe motion planning in occluded scenarios with human obstacles.

Our work has some limitations. Our augmented datasets with occlusion characteristics are synthesized from human-only action datasets. Those human actions were captured in an environment with no physical robots. The human actions in the real world in an environment shared with a robot may be different. The trajectories computed by our occlusion-aware planner may be less optimal because we may compute path detours while we first attempt to move the arms to reduce occlusion. Our overall planning algorithm uses an optimization framework with occlusion functions and is prone to local minima problems. Our motion prediction algorithm assumes that a good representation of the human skeleton can be computed from a given depth image.
There are many avenues for future work. In addition to addressing the limitations, we would like to evaluate our approach in complex scenes with multiple humans, which can result in complex occlusion relationships. 